# Discovery of non-gaussian linear causal models using ICA

**Shohei Shimizu**[*]
HIIT Basic Research Unit
Dept. of Comp. Science
University of Helsinki
Finland

**Aapo Hyvärinen**
HIIT Basic Research Unit
Dept. of Comp. Science
University of Helsinki
Finland

**Yutaka Kano**
Div. Mathematical Science
Osaka University
Japan

**Patrik O. Hoyer**
HIIT Basic Research Unit
Dept. of Comp. Science
University of Helsinki
Finland

## Abstract

In recent years, several methods have been proposed for the discovery of causal structure from non-experimental data (Spirtes et al. 2000; Pearl 2000). Such methods make various assumptions on the data generating process to facilitate its identification from purely observational data. Continuing this line of research, we show how to discover the complete causal structure of continuous-valued data, under the assumptions that (a) the data generating process is linear, (b) there are no unobserved confounders, and (c) disturbance variables have non-gaussian distributions of non-zero variances. The solution relies on the use of the statistical method known as independent component analysis (ICA), and does not require any pre-specified time-ordering of the variables. We provide a complete Matlab package for performing this LiNGAM analysis (short for Linear Non-Gaussian Acyclic Model), and demonstrate the effectiveness of the method using artificially generated data.

## 1 INTRODUCTION

Several authors (Spirtes et al. 2000; Pearl 2000) have recently formalized concepts related to causality using probability distributions defined on directed acyclic graphs. This line of research emphasizes the importance of understanding the process which generated the data, rather than only characterizing the joint distribution of the observed variables. The reasoning is that a causal understanding of the data is essential to be able to predict the consequences of interventions, such as setting a given variable to some specified value.

One of the main questions one can answer using this kind of theoretical framework is: 'Under what circumstances and in what way can one determine causal structure on the basis of observational data alone?'. In many cases it is impossible or too expensive to perform controlled experiments, and hence methods for discovering likely causal relations from uncontrolled data would be very valuable.

Existing discovery algorithms (Spirtes et al. 2000; Pearl 2000) generally work in one of two settings. In the case of discrete data, no functional form for the dependencies is usually assumed. On the other hand, when working with continuous variables, a linear-gaussian approach is almost invariably taken.

In this paper, we show that when working with continuous-valued data, a significant advantage can be achieved by departing from the gaussianity assumption. While the linear-gaussian approach usually only leads to a *set* of possible models, equivalent in their conditional correlation structure, a linear-*non-gaussian* setting allows the full causal model to be estimated, with no undetermined parameters.

The paper is structured as follows. First, in section 2, we describe our assumptions on the data generating process. These assumptions are essential for the application of our causal discovery method, detailed in sections 3 through 5. Section 6 discusses how one can test whether the found model seems plausible. In section 7 we empirically verify that our algorithm works as stated. Conclusions and future research directions are given in section 8.

## 2 LINEAR CAUSAL NETWORKS

Assume that we observe data generated from a process with the following properties:

1. The observed variables $x_i$, $i = \{1 \ldots n\}$ can be arranged in a *causal order*, such that no later vari-

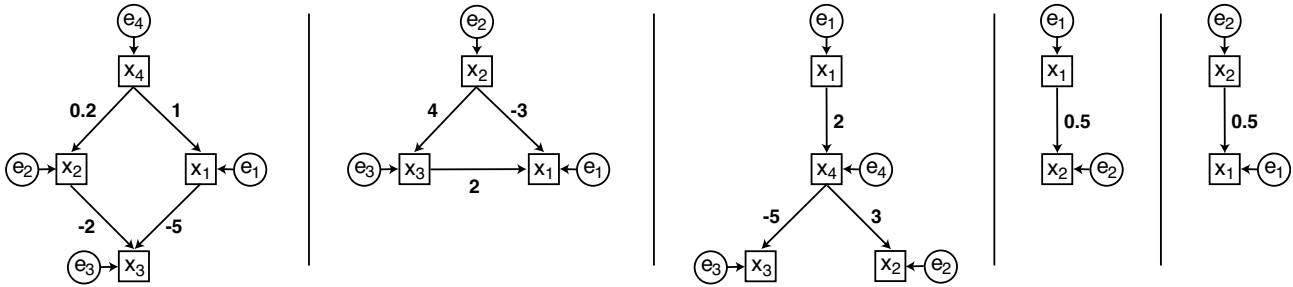

Figure 1: A few examples of data generating models satisfying our assumptions. For example, in the leftmost model, the data is generated by first drawing the $e_i$ independently from their respective non-gaussian distributions, and subsequently setting (in this order) $x_4 = e_4$, $x_2 = 0.2x_4 + e_2$, $x_1 = x_4 + e_1$, and $x_3 = -2x_2 - 5x_1 + e_3$. (Here, we have assumed for simplicity that all the $c_i$ are zero, but this may not be the case in general.) Note that the variables are not causally sorted (reflecting the fact that we usually do not know the causal ordering a priori), but that in each of the graphs they *can* be arranged in a causal order, as all graphs are directed acyclic graphs. In this paper we show that the full causal structure, including all parameters, are identifiable given a sufficient number of observed data vectors **x**.

able causes any earlier variable. We denote such a causal order by $k(i)$. That is, the generating process is *recursive* (Bollen 1989), meaning it can be represented graphically by a *directed acyclic graph* (DAG) (Pearl 2000; Spirtes et al. 2000).

2. The value assigned to each variable $x_i$ is a *linear function* of the values already assigned to the earlier variables, plus a 'disturbance' (noise) term $e_i$, and plus an optional constant term $c_i$, that is

$$x_i = \sum_{k(j)<k(i)} b_{ij} x_j + e_i + c_i. \quad (1)$$

3. The disturbances $e_i$ are all continuous random variables with *non-gaussian* distributions of non-zero variances, and the $e_i$ are independent of each other, i.e. $p(e_1, \ldots, e_n) = \prod_i p_i(e_i)$.

A model with these three properties we call a *Linear, Non-Gaussian, Acyclic Model*, abbreviated LiNGAM.

We assume that we are able to observe a large number of data vectors **x** (which contain the components $x_i$), and each is generated according to the above described process, with the same causal order $k(i)$, same coefficients $b_{ij}$, same constants $c_i$, and the disturbances $e_i$ sampled independently from the same distributions.

Note that the above assumptions imply that there are *no unobserved confounders* (Pearl 2000). Spirtes et al. (2000) call this the *causally sufficient* case. See figure 1 for a few examples of data models fulfilling the assumptions of our model.

A key difference to most earlier work on the linear, causally sufficient, case is the assumption of non-gaussianity of the disturbances. In most work, an explicit or implicit assumption of gaussianity has been made (Bollen 1989; Spirtes et al. 2000). An assumption of gaussianity of disturbance variables makes the full joint distribution over the $x_i$ gaussian, and the covariance matrix of the data embodies all one could possibly learn from observing the variables. Hence, all conditional correlations can be computed from the covariance matrix, and discovery algorithms based on conditional independence can be easily applied.

However, it turns out that an assumption of *non-gaussianity* may actually be more useful. In particular, it turns out that when this assumption is valid, the complete causal structure can in fact be estimated, without any prior information on a causal ordering of the variables. This is in stark contrast to what can be done in the gaussian case: algorithms based only on second-order statistics (i.e. the covariance matrix) are generally not able to discern the full causal structure in most cases. The simplest such case is that of two variables, $x_1$ and $x_2$. A method based only on the covariance matrix has no way of preferring $x_1 \rightarrow x_2$ over the reverse model $x_1 \leftarrow x_2$; indeed the two are indistinguishable in terms of the covariance matrix (Spirtes et al. 2000). However, assuming non-gaussianity, one can actually discover the direction of causality, as first shown by Shimizu and Kano (2003b). This result can be extended to several variables (Shimizu et al. 2005). Here, we further develop the method so as to estimate the full linear model, including all parameters.

## 3 MODEL IDENTIFICATION USING INDEPENDENT COMPONENT ANALYSIS

The key to the solution to the linear discovery problem is to realize that the observed variables are linear func-tions of the disturbance variables, and the disturbance variables are mutually independent and non-gaussian. If we as preprocessing subtract out the mean of each variable $x_i$, we are left with the following system of equations:

$$\mathbf{x} = \mathbf{Bx} + \mathbf{e}, \qquad (2)$$

where $\mathbf{B}$ is a matrix that could be permuted (by simultaneous equal row and column permutations) to strict lower triangularity if one knew a causal ordering $k(i)$ of the variables. (Strict lower triangularity is here defined as lower triangular with all zeros on the diagonal.) Solving for $\mathbf{x}$ one obtains

$$\mathbf{x} = \mathbf{Ae}, \qquad (3)$$

where $\mathbf{A} = (\mathbf{I} - \mathbf{B})^{-1}$. Again, $\mathbf{A}$ could be permuted to lower triangularity (although not *strict* lower triangularity, actually in this case all diagonal elements will be *non-zero*) with an appropriate permutation $k(i)$. Taken together, equation (3) and the independence and non-gaussianity of the components of $\mathbf{e}$ define the standard linear *independent component analysis* model.

Independent component analysis (ICA) (Comon 1994; Hyvärinen et al. 2001) is a fairly recent statistical technique for identifying a linear model such as that given in equation (3). If the observed data is a linear, invertible mixture of non-gaussian independent components, it can be shown (Comon 1994) that the mixing matrix $\mathbf{A}$ is identifiable (up to scaling and permutation of the columns, as discussed below) given enough observed data vectors $\mathbf{x}$. Furthermore, efficient algorithms for estimating the mixing matrix are available (Hyvärinen 1999).

We again want to emphasize that ICA uses non-gaussianity (that is, more than covariance information) to estimate the mixing matrix $\mathbf{A}$ (or equivalently its inverse $\mathbf{W} = \mathbf{A}^{-1}$). For gaussian disturbance variables $e_i$, ICA cannot in general find the correct mixing matrix because many different mixing matrices yield the same covariance matrix, which in turn implies the exact same gaussian joint density. Our requirement for non-gaussianity of disturbance variables stems from the same requirement in ICA.

While ICA is essentially able to estimate $\mathbf{A}$ (and $\mathbf{W}$), there are two important indeterminacies that ICA cannot solve: First and foremost, the order of the independent components is in no way defined or fixed. Thus, we could reorder the independent components and, correspondingly, the columns of $\mathbf{A}$ (and rows of $\mathbf{W}$) and get an equivalent ICA model (the same probability density for the data). In most applications of ICA, this indeterminacy is of no significance and can be ignored, but in LiNGAM, we can and we have to find the correct permutation as described in section 4 below.

The second indeterminacy of ICA concerns the scaling of the independent components. In ICA, this is usually handled by assuming all independent components to have unit variance, and scaling $\mathbf{W}$ and $\mathbf{A}$ appropriately. On the other hand, in LiNGAM (as in SEM) we allow the disturbance variables to have arbitrary (non-zero) variances, but fix their weight (connection strength) to their corresponding observed variable to unity. This requires us to re-normalize the rows of $\mathbf{W}$ so that all the diagonal elements equal unity, before computing $\mathbf{B}$, as described in the LiNGAM algorithm.

Our discovery algorithm, detailed in the next section, can be briefly summarized as follows: First, use a standard ICA algorithm to obtain an estimate of the mixing matrix $\mathbf{A}$ (or equivalently of $\mathbf{W}$), and subsequently permute it and normalize it appropriately before using it to compute $\mathbf{B}$ containing the sought connection strengths $b_{ij}$.

## 4 LiNGAM DISCOVERY ALGORITHM

Based on the observations given in sections 2 and 3, we propose the following causal discovery algorithm:

---

LiNGAM discovery algorithm

1. Given an $n \times m$ data matrix $\mathbf{X}$ ($n \ll m$), where each column contains one sample vector $\mathbf{x}$, first subtract the mean from each row of $\mathbf{X}$, then apply an ICA algorithm to obtain a decomposition $\mathbf{X} = \mathbf{AS}$ where $\mathbf{S}$ has the same size as $\mathbf{X}$ and contains in its rows the independent components. From here on, we will exclusively work with $\mathbf{W} = \mathbf{A}^{-1}$.

2. Find the one and only permutation of rows of $\mathbf{W}$ which yields a matrix $\widetilde{\mathbf{W}}$ without any zeros on the main diagonal. In practice, small estimation errors will cause all elements of $\mathbf{W}$ to be non-zero, and hence the permutation is sought which minimizes $\sum_i 1/|\widetilde{\mathbf{W}}_{ii}|$.

3. Divide each row of $\widetilde{\mathbf{W}}$ by its corresponding diagonal element, to yield a new matrix $\widetilde{\mathbf{W}}'$ with all ones on the diagonal.

4. Compute an estimate $\widehat{\mathbf{B}}$ of $\mathbf{B}$ using $\widehat{\mathbf{B}} = \mathbf{I} - \widetilde{\mathbf{W}}'$.

5. Finally, to find a causal order, find the permutation matrix $\mathbf{P}$ (applied equally to both rows and columns) of $\widehat{\mathbf{B}}$ which yields a matrix $\widetilde{\mathbf{B}} = \mathbf{P}\widehat{\mathbf{B}}\mathbf{P}^T$ which is as

> close as possible to strictly lower triangular. This can be measured for instance using $\sum_{i \leq j} \widehat{\mathbf{B}}_{ij}^2$.

A complete Matlab code package implementing this algorithm is available online at our LiNGAM homepage: http://www.cs.helsinki.fi/group/neuroinf/lingam/

We now describe each of these steps in more detail.

In the first step of the algorithm, the ICA decomposition of the data is computed. Here, any standard ICA algorithm can be used. Although our implementation uses the FastICA algorithm (Hyvärinen 1999), one could equally well use one of the many other algorithms available, see e.g. (Hyvärinen et al. 2001). However, it is important to select an algorithm which can estimate independent components of many different distributions, as in general the distributions of the disturbance variables will not be known in advance.

Because of the permutation indeterminacy of ICA, the rows of $\mathbf{W}$ will be in random order. This means that we do not yet have the correct correspondence between the disturbance variables $e_i$ and the observed variables $x_i$. The former correspond to the rows of $\mathbf{W}$ while the latter correspond to the columns of $\mathbf{W}$. Thus, our first task is to permute the rows to obtain a correspondence between the rows and columns. If $\mathbf{W}$ were estimated exactly, there would be only a single row permutation that would give a matrix with no zeros on the diagonal, and this permutation gives the correct correspondence. (A proof of this is given in Appendix A.)

In practice, however, ICA algorithms applied on finite data sets will yield estimates which are only approximately zero for those elements which should be exactly zero. Thus, our algorithm searches for the permutation using a cost function which heavily penalizes small absolute values in the diagonal, as specified in step 2. In addition to being intuitively sensible, this cost function can also be derived from a maximum-likelihood framework; for details, see Appendix B.

When the number of observed variables $x_i$ is relatively small (less than eight or so) then finding the best permutation is easy, since a simple exhaustive search can be performed. This is what our current implementation uses. For larger dimensionalities, however, this quickly becomes infeasible. We are currently developing more efficient methods to tackle cases with tens of variables or more. For up-to-date information, refer to the LiNGAM webpage (see URL above).

Once we have obtained the correct correspondence between rows and columns of the ICA decomposition, calculating our estimates of the $b_{ij}$ is straightforward. First, we normalize the rows of the permuted matrix to yield a diagonal with all ones, and then remove this diagonal and flip the sign of the remaining coefficients, as specified in steps 3 and 4.

Although we now have estimates of all coefficients $b_{ij}$ we do not yet have available a causal ordering $k(i)$ of the variables. Such an ordering (in general there may exist many if the generating network is not fully connected) is important for visualizing the resulting graph, and also plays a role in our edge pruning method. A causal ordering can be found by permuting both rows and columns (using the same permutation) of the matrix $\widehat{\mathbf{B}}$ (containing the estimated connection strengths) to yield a strictly lower triangular matrix. If the estimates were exact, this would be a trivial task. However, since our estimates will not contain exact zeros, we will have to settle for approximate strict lower triangularity, measured for instance as described in step 5. Again, finding the best permutation can be done brute-force for low dimensionalities (our current implementation), but for higher dimensionalities a more sophisticated method is required. We hope to have such a method implemented in our code package by the time you read this.

## 5  PRUNING EDGES

After finding a causal ordering $k(i)$, we can set to zero the coefficients of $\widehat{\mathbf{B}}$ which are implied zero by the order (i.e. those corresponding to the upper triangular part of the causally permuted connection matrix $\widetilde{\mathbf{B}}$). However, all remaining connections are in general non-zero. Even estimated connection strengths which are exceedingly weak (and hence probably zero in the generating model) remain and the network is fully connected. Both for achieving an intuitive understanding of the data, and especially for visualization purposes, a pruned network would be desirable.

Fortunately, pruning edges *once we know a causal ordering of the variables* is a well-known problem, and extensively discussed in the Structural Equation Modeling (SEM) tradition (Bollen 1989). Here, we propose a basic method based on this correspondence. In our implementation we take the causal ordering obtained from the LiNGAM algorithm, and then simply estimate the connection strengths using covariance information alone for different resamplings of the original data. In this way, it is possible to obtain measures of the variances of the estimates of the $b_{ij}$, and use these variances to prune those edges whose estimated means are low compared with their standard deviations. Future versions of our software package should incorporate the more advanced methods developed in the SEM community, possibly taking into account the non-gaussianity of the data as well (Shimizu and Kano 2003b).

## 6 TESTING THE ASSUMPTIONS

The LiNGAM algorithm consistently estimates the connection strengths (and a causal order) if the model assumptions hold and the amount of data is sufficient. But what if our assumptions do not in fact hold? In such a case there is of course no guarantee that the proposed discovery algorithm will find true causal relationships between the variables.

The good news is that, in some cases, it is possible to detect problems. If, for instance, the estimated matrix $\widehat{\mathbf{B}}$ cannot be permuted to yield anything close to a strictly lower triangular matrix, one knows that one or more of the assumptions do not hold and one thus cannot count on the results. Thus, our current implementation reports how well the triangularity condition holds, and warns the user if the end result is far from triangular.

Another possible test is whether the components (rows of $\mathbf{S}$) are actually independent. If the 'independent components' found by ICA are in fact not very independent, then the data did not arise from the assumed model. Independence of continuous random variables is a property which is more problematic to test than the triangularity condition. Since the linear correlations are in fact forced to zero by many ICA algorithms (including the one we use) it is of no use to test these. Rather, tests based on some form of nonlinear correlations must be used (Murata 2001; Shimizu and Kano 2003a). Although not in our current first implementation, we hope to add such tests to our code package as soon as possible.

Unfortunately, however, it is never possible to completely confirm the assumptions (and hence the found causal model) purely from observational data. Experiments, where the individual variables are explicitly manipulated (often by random assignment) and their effects monitored, are the only way to verify any causal model. Nevertheless, it is fair to say that if both the triangularity and the independence conditions hold then the found causal model is likely to hold, at least approximately. Only pathological cases constructed by mischievous data designers seem likely to be problematic for our framework. Thus, we think that a LiNGAM analysis will prove a useful first step in many cases for providing educated guesses of causal models, which might subsequently be verified in systematic experiments.

## 7 EXPERIMENTS

To verify the validity of our method (and of our Matlab code), we performed extensive experiments with simulated data. All experimental code (including the precise code to produce figure 2) is included in the LiNGAM code package.

We repeatedly performed the following experiment:

1. First, we randomly constructed a strictly lower-triangular matrix $\mathbf{B}$. Various dimensionalities (8 or less) were used. Both fully connected (no zeros in the strictly lower triangular part) and sparse networks (many zeros) were tested. We also randomly selected variances of the disturbance variables and values for the constants $c_i$.

2. Next, we generated data by independently drawing the disturbance variables $e_i$ from gaussian distributions and subsequently passing them through a power non-linearity (raising the absolute value to an exponent in the interval [0.5, 0.8] or [1.2, 2.0], but keeping the original sign) to make them non-gaussian. Various data set sizes were tested. The $e_i$ were then scaled to yield the desired variances, and the observed data $\mathbf{X}$ was generated according to the assumed recursive process.

3. Before feeding the data to the LiNGAM algorithm, we randomly permuted the rows of the data matrix $\mathbf{X}$ to hide the causal order with which the data was generated. At this point, we also permuted $\mathbf{B}$, the $c_i$, as well as the variances of the disturbance variables to match the new order in the data.

4. Finally, we fed the data to our discovery algorithm, and compared the estimated parameters to the generating parameters. In particular, we made a scatterplot of the entries in the estimated matrix $\widehat{\mathbf{B}}$ against the corresponding ones in $\mathbf{B}$.

Since the number of different possible parameter configurations is limitless, we feel that the reader is best convinced by personally running the simulations using various settings. This can be easily done by anyone with access to Matlab.[1] Nevertheless, we here show some representative results.

Figure 2 gives combined scatterplots of the elements of $\widehat{\mathbf{B}}$ versus those of $\mathbf{B}$. The different plots correspond to different dimensionalities (numbers of variables) and different data sizes (numbers of data vectors), where each plot combines the data for a number of different network sparseness levels and non-linearities. Although for very small data sizes the estimation often fails, when the data size grows the estimation works practically flawlessly, as evidenced by the grouping of the datapoints onto the main diagonal.

---

[1] Note that a version running on the freely available Octave software is currently in development and might be available by the time you read this.

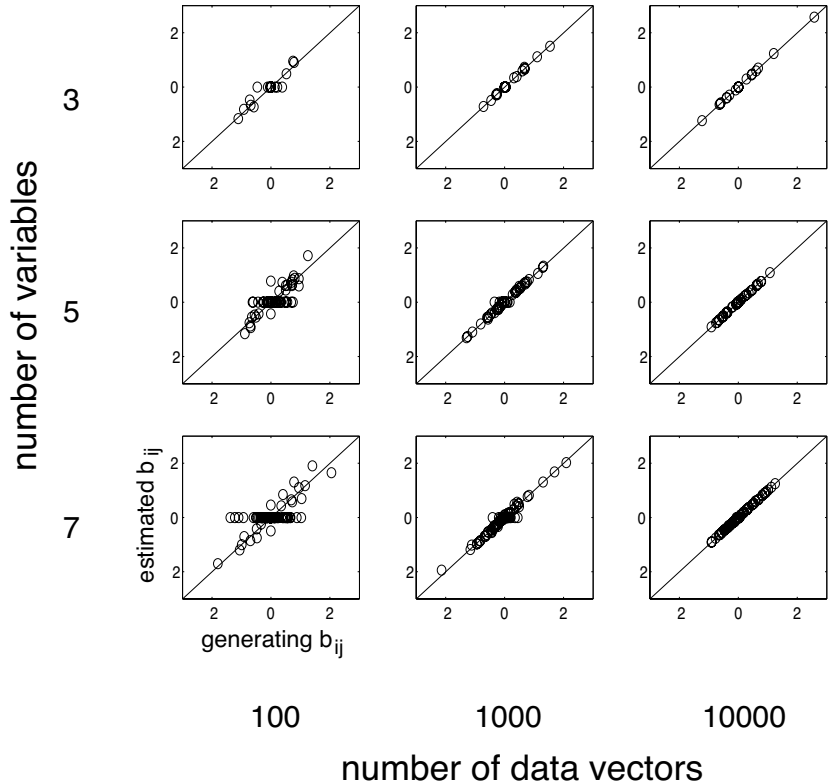

Figure 2: Scatterplots of the estimated $b_{ij}$ versus the original (generating) values. The different plots correspond to different numbers of variables and different numbers of data vectors. Although for small data sizes the estimation often fails, when there is sufficient data the estimation works essentially flawlessly, as evidenced by the grouping of the points along the diagonal.

In summary, the experiments verify the correctness of the method, and demonstrate that reliable estimation is possible even with fairly limited amounts of data.

## 8 CONCLUSIONS AND FUTURE WORK

Developing methods for causal inference from non-experimental data is a fundamental problem with a very large number of potential applications. Although one can never fully prove the validity of a causal model from observational data alone, such methods are nevertheless crucial in cases where it is impossible or very costly to perform experiments.

Previous methods developed for linear causal models (Bollen 1989; Spirtes et al. 2000; Pearl 2000) have been based on an explicit or implicit assumption of gaussianity, and have hence been based solely on the covariance structure of the data. Because of this, additional information (such as the time-order of the variables) is usually required to obtain a full causal model of the variables. Without such information, algorithms based on the gaussianity assumption cannot in most cases distinguish between multiple equally possible causal models.

In this paper, we have shown that an assumption of non-gaussianity of the disturbance variables, together with the assumption of linearity and causal sufficiency, allows the causal model to be completely identified. Furthermore, we have provided a practical algorithm which estimates the causal structure under these assumptions.

Future work will focus on implementational issues in problems involving tens of variables or more. It remains an open question what algorithms are best suited for optimizing our objective functions, which measure the goodness of the permutations, in such cases. Further, the practical value of the LiNGAM analysis needs to be determined by applying it to real-world datasets and comparing it to other methods for causal inference from non-experimental data. In many cases involving real-world data, practitioners in the field already have a fairly good understanding of the causal processes underlying the data. An interesting question is how well methods such as ours do on such datasets. For the most recent developments, please see the webpage:

http://www.cs.helsinki.fi/group/neuroinf/lingam/

**Acknowledgements**

The authors would like to thank Alex Pothen and Heikki Mannila for discussions relating to algorithms for solving the permutation problems. S.S. was supported by Grant-in-Aid for Scientific Research from the Ministry of Education, Culture and Sports, Japan. A.H. was supported by the Academy of Finland through an Academy Research Fellow Position and project #203344. P.O.H. was supported by the Academy of Finland project #204826.

## A PROOF OF UNIQUENESS OF ROW PERMUTATION

Here, we show that, were the estimates of ICA exact, there is only a single permutation of the rows of $\mathbf{W}$ which results in a diagonal with no zero entries.

It is well-known (Bollen 1989) that the DAG structure of the network guarantees that for some permutation of the variables, the matrix $\mathbf{B}$ is strictly lower-triangular. This implies that the correct $\widetilde{\mathbf{W}}$ (where the disturbance variables are aligned with the observed variables) can be permuted to lower-triangular form (with no zero entries on the diagonal) by equal row and column permutations, i.e.

$$\widetilde{\mathbf{W}} = \mathbf{P}_d \mathbf{M} \mathbf{P}_d^T, \qquad (4)$$

where $\mathbf{M}$ is lower-triangular and has no zero entries on the diagonal, and $\mathbf{P}_d$ is a permutation matrix representing a causal ordering of the variables. Now, ICA returns a matrix with randomly permuted rows,

$$\mathbf{W} = \mathbf{P}_{\text{ica}} \widetilde{\mathbf{W}} = \mathbf{P}_{\text{ica}} \mathbf{P}_d \mathbf{M} \mathbf{P}_d^T = \mathbf{P}_1 \mathbf{M} \mathbf{P}_2^T, \quad (5)$$

where $\mathbf{P}_{\text{ica}}$ is the random ICA row permutation, and on the right we have denoted by $\mathbf{P}_1 = \mathbf{P}_{\text{ica}} \mathbf{P}_d$ and $\mathbf{P}_2 = \mathbf{P}_d$, respectively, the row and column permutations from the lower triangular matrix $\mathbf{M}$.

We now prove that $\mathbf{W}$ has no zero entries on the diagonal if and only if the row and column permutations are equal, i.e. $\mathbf{P}_1 = \mathbf{P}_2$. Hence, there is only one row permutation of $\mathbf{W}$ which yields no zero entries on the diagonal, and it is the one which finds the correspondence between the disturbance variables and the observed variables.

**Lemma 1** *Assume $\mathbf{M}$ is lower triangular and all diagonal elements are nonzero. A permutation of rows and columns of $\mathbf{M}$ has only non-zero entries in the diagonal if and only if the row and column permutations are equal.*

Proof: First, we prove that if the row and columns permutations are not equal, there will be zero elements in the diagonal.

Denote by $\mathbf{K}$ a lower triangular matrix of all ones in the lower triangular part. Denote by $\mathbf{P}_1$ and $\mathbf{P}_2$ two permutation matrices. The number of non-zero diagonal entries in a permuted version of $\mathbf{K}$ is $\text{tr}(\mathbf{P}_1 \mathbf{K} \mathbf{P}_2^T)$. This is the maximum number of non-zero diagonal entries when an arbitrary lower triangular matrix is permuted.

We have $\text{tr}(\mathbf{P}_1\mathbf{K}\mathbf{P}_2^T) = \text{tr}(\mathbf{K}\mathbf{P}_2^T\mathbf{P}_1)$. Thus, we first consider permutations of columns only, given by $\mathbf{P}_2^T\mathbf{P}_1$. Assume the columns of $\mathbf{K}$ are permuted so that the permutation is not equal to identity. Then, there exists an index $i$ so that the column of index $i$ has been moved to column index $j$ where $j < i$ (If there were no such columns, all the columns would be moved to the right, which is impossible.) Obviously, the diagonal entry in the $j$-th column in the permuted matrix is zero. Thus, any column permutation not equal to the identity creates at least one zero entry in the diagonal.

Thus, to have nonzero diagonal, we must have $\mathbf{P}_2^T\mathbf{P}_1 = \mathbf{I}$. This means that the column and row permutations must be equal.

Next, assume that the row and column permutations are equal. Consider $\mathbf{M} = \mathbf{I}$ as a worst-case scenario. Then the permuted matrix equals $\mathbf{P}_1\mathbf{I}\mathbf{P}_2^T$ which equals identity, and all the diagonal elements are nonzero. Adding more nonzero elements in the matrix only increases the number of nonzero elements in the permuted version.

Thus, the lemma is proven.

## B ML DERIVATION OF OBJECTIVE FUNCTION FOR FINDING THE CORRECT ROW PERMUTATION

Since the ICA estimates are never exact, all elements of $\mathbf{W}$ will be non-zero, and one cannot base the permutation on exact zeros. Here we show that the objective function for step 2 of the LiNGAM algorithm can be derived from a maximum likelihood framework.

Let us denote by $e_{it}$ the value of disturbance variable $i$ for the $t$:th datavector of the dataset. Assume that we model the disturbance variables $e_{it}$ by a generalized gaussian density:

$$\log p(e_{it}) = -|e_{it}|^\alpha/\beta + Z \qquad (6)$$

where the $\alpha, \beta$ are parameters and $Z$ is a normalization constant. Then, the log-likelihood of the model equals

$$\sum_t \sum_i -\left|\frac{e_{it}}{\beta w_{ii}}\right|^\alpha = -\sum_i \frac{1}{\beta|w_{ii}|^\alpha}\sum_t |e_{it}|^\alpha \qquad (7)$$

because each row of $\mathbf{W}$ is subsequently divided by its diagonal element. To maximize the likelihood, we find the permutation of rows for which the diagonal elements maximize this term. For simplicity, assuming that the pdf's of all independent components are the same, this means we solve

$$\min_{\text{all row perms}} \sum_i \frac{1}{|w_{ii}|^\alpha} \qquad (8)$$

In principle, we could estimate $\alpha$ from the data using ML estimation as well, but for simplicity we fix it to unity because it does not really change the qualitative behaviour of the objective function. Regardless of its value, this objective function heavily penalizes small values on the diagonal, as we intuitively (based on the argumentation in section 4) require.